\documentclass[acmtog]{acmart}

\usepackage{booktabs}       %
\usepackage{amsfonts}       %
\usepackage{nicefrac}       %
\usepackage{microtype}      %
\usepackage{xcolor}         %
\usepackage[capitalize]{cleveref}
\usepackage{graphicx}
\usepackage{tabularx}
\usepackage{multirow}

\usepackage[ruled]{algorithm2e} %

\SetAlFnt{\small}
\SetAlCapFnt{\small}
\SetAlCapNameFnt{\small}
\SetAlCapHSkip{0pt}
\IncMargin{-\parindent}

\acmJournal{TOG}
\acmVolume{9}
\acmNumber{4}
\acmArticle{39}
\acmYear{2010}
\acmMonth{3}
\acmSubmissionID{249}

\copyrightyear{2023}
\acmYear{2023}
\setcopyright{acmlicensed}\acmConference[SA Conference Papers '23]{SIGGRAPH Asia 2023 Conference Papers}{December 12--15, 2023}{Sydney, NSW, Australia}
\acmBooktitle{SIGGRAPH Asia 2023 Conference Papers (SA Conference Papers '23), December 12--15, 2023, Sydney, NSW, Australia}
\acmPrice{15.00}
\acmDOI{10.1145/3610548.3618149}
\acmISBN{979-8-4007-0315-7/23/12}

\setcitestyle{square}

\begin{document}
\title{\invs: Repurposing Diffusion Inpainters for Novel View Synthesis}

\author{Yash Kant}
\orcid{0009-0002-8347-4895}
\email{ysh.kant@gmail.com}
\affiliation{%
  \institution{University of Toronto}
  \city{Toronto}
  \country{Canada}
}
\affiliation{%
  \institution{Snap Research}
  \city{Toronto}
  \country{Canada}
}

\author{Aliaksandr Siarohin}
\email{asiarohin@snapchat.com}
\affiliation{%
  \institution{Snap Research}
  \city{Los Angeles}
  \country{USA}
}

\author{Michael Vasilkovsky}
\email{mvasilkovsky@snapchat.com}
\affiliation{%
  \institution{Snap Research}
  \city{Los Angeles}
  \country{USA}
}

\author{Riza Alp Guler}
\email{rguler@snap.com}
\affiliation{%
  \institution{Snap Research}
  \city{London}
  \country{UK}
}

\author{Jian Ren}
\email{jren@snap.com}
\affiliation{%
  \institution{Snap Research}
  \city{Los Angeles}
  \country{USA}
}

\author{Sergey Tulyakov}
\email{stulyakov@snap.com}
\affiliation{%
  \institution{Snap Research}
  \city{Los Angeles}
  \country{USA}
}

\author{Igor Gilitschenski}
\email{gilitschenski@cs.toronto.edu}
\affiliation{%
  \institution{University of Toronto}
  \city{Toronto}
  \country{Canada}
}

\newcommand{\new}[1]{{#1}}

\crefname{section}{Sec.}{Secs.}
\Crefname{section}{Section}{Sections}
\crefname{table}{Tab.}{Tabs.}
\Crefname{table}{Table}{Tables}
\crefname{figure}{Fig.}{Figs.}
\Crefname{figure}{Figure}{Figures}
\crefname{equation}{Eq.}{Eqs.}
\Crefname{equation}{Equation}{Equations}
\hyphenpenalty=1200

\newcommand{\SD}{\emph{SD}\xspace}
\newcommand{\ISD}{\emph{ISD}\xspace}
\newcommand{\MDP}{\emph{MDP}\xspace}
\newcommand{\objaverse}{\emph{Objaverse}\xspace}
\newcommand{\pointe}{\emph{Point-E}\xspace}
\newcommand{\shapee}{\emph{Shap-E}\xspace}
\newcommand{\zeroone}{\emph{Zero-1-to-3}\xspace}
\newcommand{\gso}{\emph{GSO}\xspace}
\newcommand{\rtmv}{\emph{RTMV}\xspace}
\newcommand{\cothreed}{\emph{CO3D}\xspace}
\newcommand{\invs}{\emph{iNVS}\xspace}

\newcommand{\height}{h}
\newcommand{\width}{w}
\newcommand{\aka}{\textit{a.k.a.}\xspace}

\newcommand{\sourceview}{\mathbf{I}_{s}}
\newcommand{\sourcedepth}{\mathbf{D}_{s}}

\newcommand{\rotationsource}{\mathbf{R}_{s}}
\newcommand{\translationsource}{\mathbf{T}_{s}}
\newcommand{\camerasourcex}{[\mathbf{R}_{s}|\mathbf{T}_{s}]}
\newcommand{\camerasource}{\mathbf{C}_{s}}
\newcommand{\intrinsicsource}{\mathbf{K}_{s}}

\newcommand{\targetview}{\mathbf{I}_{t}}
\newcommand{\targetdepth}{\mathbf{D}_{t}}

\newcommand{\rotationtarget}{\mathbf{R}_{t}}
\newcommand{\translationtarget}{\mathbf{T}_{t}}
\newcommand{\cameratargetx}{[\mathbf{R}_{t}|\mathbf{T}_{t}]}
\newcommand{\cameratarget}{\mathbf{C}_{t}}
\newcommand{\intrinsictarget}{\mathbf{K}_t}

\newcommand{\targetsplat}{\mathbf{I}_{s \rightarrow t}}
\newcommand{\masksplat}{\mathbf{M}_{s \rightarrow t}}

\newcommand{\denoiser}{\epsilon_\theta}
\newcommand{\depthpredictor}{m_d}

\newcommand{\pointsource}{\mathbf{p}_s}
\newcommand{\pointtarget}{\mathbf{p}_t}
\newcommand{\pointworld}{\mathbf{p}_w}

\newcommand{\linet}{l_T}
\newcommand{\depth}{d_s}
\newcommand{\tdepth}{d_t}

\newcommand{\textconditioning}{X_s}

\newcommand{\real}{\mathbb{R}}

\newcommand{\ba}{\mathbf{a}}\newcommand{\bA}{\mathbf{A}}
\newcommand{\bb}{\mathbf{b}}\newcommand{\bB}{\mathbf{B}}
\newcommand{\bc}{\mathbf{c}}\newcommand{\bC}{\mathbf{C}}
\newcommand{\bd}{\mathbf{d}}\newcommand{\bD}{\mathbf{D}}
\newcommand{\be}{\mathbf{e}}\newcommand{\bE}{\mathbf{E}}
\newcommand{\bff}{\mathbf{f}}\newcommand{\bF}{\mathbf{F}} %
\newcommand{\bg}{\mathbf{g}}\newcommand{\bG}{\mathbf{G}}
\newcommand{\bh}{\mathbf{h}}\newcommand{\bH}{\mathbf{H}}
\newcommand{\bi}{\mathbf{i}}\newcommand{\bI}{\mathbf{I}}
\newcommand{\bj}{\mathbf{j}}\newcommand{\bJ}{\mathbf{J}}
\newcommand{\bk}{\mathbf{k}}\newcommand{\bK}{\mathbf{K}}
\newcommand{\bl}{\mathbf{l}}\newcommand{\bL}{\mathbf{L}}
\newcommand{\bm}{\mathbf{m}}\newcommand{\bM}{\mathbf{M}}
\newcommand{\bn}{\mathbf{n}}\newcommand{\bN}{\mathbf{N}}
\newcommand{\bo}{\mathbf{o}}\newcommand{\bO}{\mathbf{O}}
\newcommand{\bp}{\mathbf{p}}\newcommand{\bP}{\mathbf{P}}
\newcommand{\bq}{\mathbf{q}}\newcommand{\bQ}{\mathbf{Q}}
\newcommand{\br}{\mathbf{r}}\newcommand{\bR}{\mathbf{R}}
\newcommand{\bs}{\mathbf{s}}\newcommand{\bS}{\mathbf{S}}
\newcommand{\bt}{\mathbf{t}}\newcommand{\bT}{\mathbf{T}}
\newcommand{\bu}{\mathbf{u}}\newcommand{\bU}{\mathbf{U}}
\newcommand{\bv}{\mathbf{v}}\newcommand{\bV}{\mathbf{V}}
\newcommand{\bw}{\mathbf{w}}\newcommand{\bW}{\mathbf{W}}
\newcommand{\bx}{\mathbf{x}}\newcommand{\bX}{\mathbf{X}}
\newcommand{\by}{\mathbf{y}}\newcommand{\bY}{\mathbf{Y}}
\newcommand{\bz}{\mathbf{z}}\newcommand{\bZ}{\mathbf{Z}}

\newcommand{\balpha}{\boldsymbol{\alpha}}\newcommand{\bAlpha}{\boldsymbol{\Alpha}}
\newcommand{\bbeta}{\boldsymbol{\beta}}\newcommand{\bBeta}{\boldsymbol{\Beta}}
\newcommand{\bgamma}{\boldsymbol{\gamma}}\newcommand{\bGamma}{\boldsymbol{\Gamma}}
\newcommand{\bdelta}{\boldsymbol{\delta}}\newcommand{\bDelta}{\boldsymbol{\Delta}}
\newcommand{\bepsilon}{\boldsymbol{\epsilon}}\newcommand{\bEpsilon}{\boldsymbol{\Epsilon}}
\newcommand{\bzeta}{\boldsymbol{\zeta}}\newcommand{\bZeta}{\boldsymbol{\Zeta}}
\newcommand{\beeta}{\boldsymbol{\eta}}\newcommand{\bEta}{\boldsymbol{\Eta}} %
\newcommand{\btheta}{\boldsymbol{\theta}}\newcommand{\bTheta}{\boldsymbol{\Theta}}
\newcommand{\biota}{\boldsymbol{\iota}}\newcommand{\bIota}{\boldsymbol{\Iota}}
\newcommand{\bkappa}{\boldsymbol{\kappa}}\newcommand{\bKappa}{\boldsymbol{\Kappa}}
\newcommand{\blambda}{\boldsymbol{\lambda}}\newcommand{\bLambda}{\boldsymbol{\Lambda}}
\newcommand{\bmu}{\boldsymbol{\mu}}\newcommand{\bMu}{\boldsymbol{\Mu}}
\newcommand{\bnu}{\boldsymbol{\nu}}\newcommand{\bNu}{\boldsymbol{\Nu}}
\newcommand{\bxi}{\boldsymbol{\xi}}\newcommand{\bXi}{\boldsymbol{\Xi}}
\newcommand{\bomikron}{\boldsymbol{\omikron}}\newcommand{\bOmikron}{\boldsymbol{\Omikron}}
\newcommand{\bpi}{\boldsymbol{\pi}}\newcommand{\bPi}{\boldsymbol{\Pi}}
\newcommand{\brho}{\boldsymbol{\rho}}\newcommand{\bRho}{\boldsymbol{\Rho}}
\newcommand{\bsigma}{\boldsymbol{\sigma}}\newcommand{\bSigma}{\boldsymbol{\Sigma}}
\newcommand{\btau}{\boldsymbol{\tau}}\newcommand{\bTau}{\boldsymbol{\Tau}}
\newcommand{\bypsilon}{\boldsymbol{\ypsilon}}\newcommand{\bYpsilon}{\boldsymbol{\Ypsilon}}
\newcommand{\bphi}{\boldsymbol{\phi}}\newcommand{\bPhi}{\boldsymbol{\Phi}}
\newcommand{\bchi}{\boldsymbol{\chi}}\newcommand{\bChi}{\boldsymbol{\Chi}}
\newcommand{\bpsi}{\boldsymbol{\psi}}\newcommand{\bPsi}{\boldsymbol{\Psi}}
\newcommand{\bomega}{\boldsymbol{\omega}}\newcommand{\bOmega}{\boldsymbol{\Omega}}

\newcommand{\nA}{\mathbb{A}}
\newcommand{\nB}{\mathbb{B}}
\newcommand{\nC}{\mathbb{C}}
\newcommand{\nD}{\mathbb{D}}
\newcommand{\nE}{\mathbb{E}}
\newcommand{\nF}{\mathbb{F}}
\newcommand{\nG}{\mathbb{G}}
\newcommand{\nH}{\mathbb{H}}
\newcommand{\nI}{\mathbb{I}}
\newcommand{\nJ}{\mathbb{J}}
\newcommand{\nK}{\mathbb{K}}
\newcommand{\nL}{\mathbb{L}}
\newcommand{\nM}{\mathbb{M}}
\newcommand{\nN}{\mathbb{N}}
\newcommand{\nO}{\mathbb{O}}
\newcommand{\nP}{\mathbb{P}}
\newcommand{\nQ}{\mathbb{Q}}
\newcommand{\nR}{\mathbb{R}}
\newcommand{\nS}{\mathbb{S}}
\newcommand{\nT}{\mathbb{T}}
\newcommand{\nU}{\mathbb{U}}
\newcommand{\nV}{\mathbb{V}}
\newcommand{\nW}{\mathbb{W}}
\newcommand{\nX}{\mathbb{X}}
\newcommand{\nY}{\mathbb{Y}}
\newcommand{\nZ}{\mathbb{Z}}

\newcommand{\cA}{\mathcal{A}}
\newcommand{\cB}{\mathcal{B}}
\newcommand{\cC}{\mathcal{C}}
\newcommand{\cD}{\mathcal{D}}
\newcommand{\cE}{\mathcal{E}}
\newcommand{\cF}{\mathcal{F}}
\newcommand{\cG}{\mathcal{G}}
\newcommand{\cH}{\mathcal{H}}
\newcommand{\cI}{\mathcal{I}}
\newcommand{\cJ}{\mathcal{J}}
\newcommand{\cK}{\mathcal{K}}
\newcommand{\cL}{\mathcal{L}}
\newcommand{\cM}{\mathcal{M}}
\newcommand{\cN}{\mathcal{N}}
\newcommand{\cO}{\mathcal{O}}
\newcommand{\cP}{\mathcal{P}}
\newcommand{\cQ}{\mathcal{Q}}
\newcommand{\cR}{\mathcal{R}}
\newcommand{\cS}{\mathcal{S}}
\newcommand{\cT}{\mathcal{T}}
\newcommand{\cU}{\mathcal{U}}
\newcommand{\cV}{\mathcal{V}}
\newcommand{\cW}{\mathcal{W}}
\newcommand{\cX}{\mathcal{X}}
\newcommand{\cY}{\mathcal{Y}}
\newcommand{\cZ}{\mathcal{Z}}

\newcommand{\figref}[1]{Fig.~\ref{#1}}
\newcommand{\secref}[1]{Section~\ref{#1}}
\newcommand{\algref}[1]{Algorithm~\ref{#1}}
\newcommand{\eqnref}[1]{Eq.~\eqref{#1}}
\newcommand{\tabref}[1]{Tab.~\ref{#1}}

\newcommand{\snarf}{SNARF}
\newcommand{\cadex}{CaDeX}

\def\mc{\mathcal}
\def\mb{\mathbf}

\newcommand{\T}{^{\raisemath{-1pt}{\mathsf{T}}}}

\renewcommand{\Perp}{\perp\!\!\! \perp}

\makeatletter
\DeclareRobustCommand\onedot{\futurelet\@let@token\@onedot}
\def\@onedot{\ifx\@let@token.\else.\null\fi\xspace}
\def\eg{e.g\onedot} \def\Eg{E.g\onedot}
\def\ie{i.e\onedot} \def\Ie{I.e\onedot}
\def\cf{cf\onedot} \def\Cf{Cf\onedot}
\def\etc{etc\onedot}
\def\vs{vs\onedot}
\def\wrt{wrt\onedot}
\def\dof{d.o.f\onedot}
\def\etal{et~al\onedot}
\def\iid{i.i.d\onedot}
\makeatother

\renewcommand\UrlFont{\color{blue}\rmfamily}

\newcommand*\rot{\rotatebox{90}}

\definecolor{turquoise}{cmyk}{0.65,0,0.1,0.3}
\definecolor{purple}{rgb}{0.65,0,0.65}
\definecolor{dark_green}{rgb}{0, 0.5, 0}
\definecolor{orange}{rgb}{0.8, 0.6, 0.2}
\definecolor{red}{rgb}{0.8, 0.2, 0.2}
\definecolor{darkred}{rgb}{0.6, 0.1, 0.05}
\definecolor{blueish}{rgb}{0.0, 0.3, .6}
\definecolor{light_gray}{rgb}{0.7, 0.7, .7}
\definecolor{pink}{rgb}{1, 0, 1}
\definecolor{greyblue}{rgb}{0.25, 0.25, 1}

\newif\ifshowcomments
\showcommentstrue %

\newcommand{\todo}[1]{\noindent{\color{red}{\bf TODO:} {#1}}}

\ifshowcomments

    \newcommand{\oh}[1]{ \noindent {\color{red} {\bf OH:} {#1}} }
    \newcommand{\ag}[1]{ \noindent {\color{red} {\bf AG:} {#1}} }
    \newcommand{\mjb}[1]{ \noindent {\color{red} {\bf MJB:} {#1}} }
    \newcommand{\jy}[1]{ \noindent {\color{red} {\bf JY:} {#1}} }
    \newcommand{\xc}[1]{ \noindent {\color{red} {\bf XC:} {#1}} }
 \else
    \newcommand{\oh}[1]{\unskip}
    \newcommand{\ag}[1]{\unskip}
    \newcommand{\mjb}[1]{\unskip}
    \newcommand{\jy}[1]{\unskip}
    \newcommand{\xc}[1]{\unskip}
\fi   

\newcommand{\update}[1]{{\color{black}{#1}}}

\newcommand{\methodname}{gDNA\xspace}
\newcommand{\suppmat}{Sup.~Mat.\xspace}

\newcommand{\rulesep}{\unskip\ \vrule\ }

\renewcommand{\shortauthors}{Kant et al.}
\begin{abstract}
We present a method for generating consistent novel views from a single source image. Our approach focuses on maximizing the reuse of visible pixels from the source image. To achieve this, we use a monocular depth estimator that transfers visible pixels from the source view to the target view. Starting from a pre-trained 2D inpainting diffusion model, we train our method on the large-scale \objaverse dataset to learn 3D object priors. While training we use a novel masking mechanism based on epipolar lines to further improve the quality of our approach. This allows our framework to perform zero-shot novel view synthesis on a variety of objects. We evaluate the zero-shot abilities of our framework on three challenging datasets: Google Scanned Objects, Ray Traced Multiview, and Common Objects in 3D. See our webpage for more details: \url{https://yashkant.github.io/invs/}

\end{abstract}

\ccsdesc[500]{Computing methodologies~Computer vision}
\ccsdesc[300]{Computing methodologies~Computer graphics}

\keywords{ Novel View Synthesis, Diffusion Models, Inpainting}

\begin{teaserfigure}
    \includegraphics[width=\linewidth]{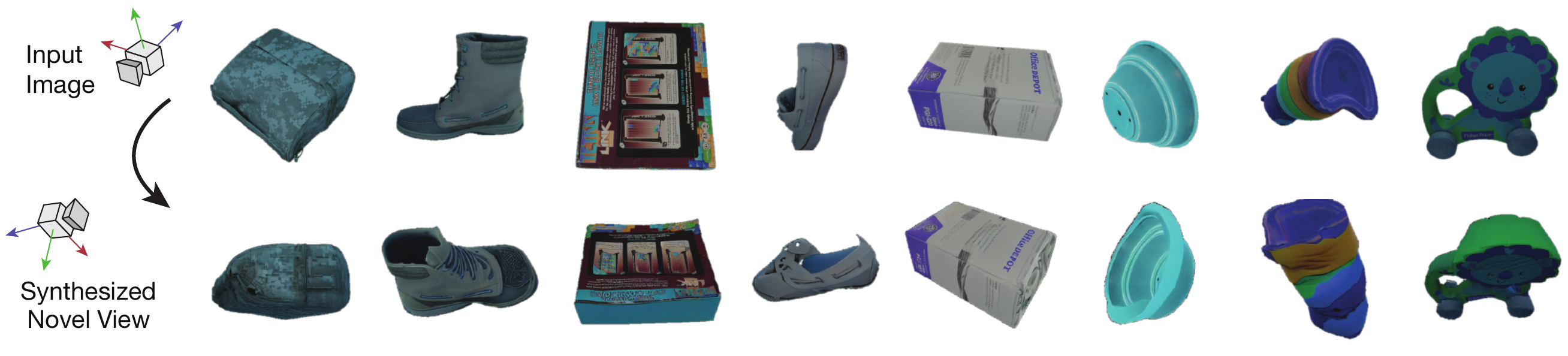}
    \caption{\textbf{Novel view synthesis results for \emph{unseen} objects.} Our system synthesizes novel view from a single image for unseen objects. We obtain detailed generations, while respecting the appearance of the region that is visible in the input image by maximizing reuse of source pixels.}
    \label{fig:teaser}
\end{teaserfigure}
\maketitle
\section{Introduction}
Synthesizing novel views from a \emph{single} image has broad applications in the computer vision fields, including object reconstruction for Augmented and Virtual Reality~\cite{choi2019extreme,tucker2020single,gao2021dynamic}, animating humans in the game and movie industry~\cite{bhatnagar2019multi,hu20213dbodynet}, and environment understanding in robotics~\cite{hu2021worldsheet,synsin}, \emph{etc}. However, generating high-fidelity novel views from one image is a longstanding challenging problem, as the image generation algorithm needs to infer the correct geometry from a partial observation of an object. Existing literature still struggles to build the generic framework to reconstruct high-quality 3D objects for image rendering. For instance, some studies require a manual 3D prior, such as human bodies~\cite{liao2023high,xiu2023econ} and faces~\cite{kemelmacher20103d,jiang20183d} for reconstruction, limiting the generation to specific categories. Other approaches suffer from inadequate reconstruction quality when dealing with single images~\cite{viewinterpolation,modelingandrendering,contextawaredepth} and often require multiple views as input~\cite{NeRF,martin2021nerf}, which may not always be available.

To tackle such a problem, we first step back to understand why we, as humans, can understand how objects should look in any view direction by just viewing the object from a single view. The reason might be that the 3D prior knowledge equipped by humans is obtained by observing a huge number of objects from various viewpoints in the real world. Thus, leveraging the knowledge learned from different objects can be a solution for accurately reconstructing objects. Recently, the emerging efforts on large-scale text-to-image diffusion models~\cite{stable_diffusion,dalle,DALLE-2,imagen} prove the capability of learning a generic object prior by training on large-scale image datasets, \emph{e.g.}, LAION~\cite{LAION}. However, these models operate in the 2D domain and lack precise control over camera view directions, limiting their effectiveness in view synthesis tasks.

This work empowers the pre-trained large-scale text-to-image diffusion model with the ability for camera viewpoint control to generate novel views.  We make the following contributions: \new{first}, we attempt to reuse pixels from the input view when camera views are not significantly far away. This is achieved by back-projecting such pixels into the 3D space using monocular depth and reprojecting them back onto the novel view. Second, we apply inpainting to recover the missing regions by leveraging Inpainting Stable Diffusion (\ISD)~\cite{stable_diffusion}. However, naïvely applying {\ISD} fails to generalize well to the masks that arise from the reprojection procedure since the {\ISD} model is trained with masks that randomly cover a part of the image. Therefore, we propose to train {\ISD} on a dataset in which we can compute such masks easily. One prominent choice is a dataset of 3D assets, \emph{e.g.}, \objaverse~\cite{deitke2022objaverse}, which can be rendered from multiple views and for which such masks can be computed. After training, our method can predict missing pixels in the novel view image, while at the same time preserving pixels that are initially visible. \new{We abbreviate our method as {\invs} which stands for inpainting-driven Novel View Synthesis.}

\new{We conduct experiments on synthetic and real datasets, and find that our method can achieve strong novel view synthesis (NVS) results from single images as shown in Figure~\ref{fig:teaser}. We conduct ablative and failure mode analyses which demonstrates that a good monocular depth estimator is important to preserve structure and allow maximal reuse of source pixels.}

\section{Related works}
\textbf{Novel View Synthesis in Space.} Novel view synthesis is a long-standing problem in computer vision and graphics. Early methods rely on the images from multiple viewpoints and attempt to incorporate the knowledge from epipolar geometry to perform smooth interpolation between the different views~\cite{viewinterpolation,modelingandrendering}. One of the important milestones in novel view synthesis is the introduction of Neural Radiance Fields (NeRFs)~\cite{NeRF}. NeRFs can synthesize smooth interpolations between different views with the help of volumetric rendering. Since then, numerous improvements have been introduced to improve the original design~\cite{NeRF--, NeRF++, NeRV, NeROIC, mip-NeRF}. However, most of them share the same limitation of relying on multiple views for learning 3D representation. 

Newer works have demonstrated that using deep networks is a promising approach for synthesizing novel views from few images owing to their generalization capabilities~\cite{sajjadi2022scene,chan2023generative,deng20233d,zhou2023sparsefusion, mirzaei2023reference}. In its limit, this approach allows for generating novel views given exactly a \emph{single} image~\cite{contextawaredepth,synsin,gu2023nerfdiff,tang2023make,shen2023anything3d} and we adapt this setting in our work. Zero123~\cite{zero1to3} proposes to fine-tune Stable Diffusion for NVS task. They condition diffusion both on the source image and on the CLIP embedding of the source image. However, this method largely ignores the inability of U-Net~\cite{unet} networks to generate output that is not aligned with the source~\cite{MonkeyNet, FOMM}. In contrast, our method relies on geometry clues to align the source and target views. This helps us to preserve the content from the source image well.

\textbf{Novel View Synthesis in Time.} Video generation conditional on one or more input image(s) can be seen as a novel view synthesis task with generated images unrolling both in space and time. Prior works~\cite{vondrick2017generating,wang2017predrnn,villegas2017decomposing,hsieh2018learning,finn2016unsupervised,denton2018stochastic} in this domain proposed the use of spatiotemporal conditioning to handle dynamic scenes. More recent works have trained on large indoor scene and video datasets further improving quality of generations task~\cite{koh2021pathdreamer,lee2021revisiting,ye2019compositional,yu2022generating}.  

Our work is particularly inspired by InfiniteNature and InfiniteNature Zero works~\cite{infinite_nature,li2022infinitenature} which utilize softmax-splatting~\cite{niklaus2020softmax} for synthesizing infinite videos of nature with a fly-through camera. Recent works in this space have tackled generation of novel views with full $360$-degree camera control~\cite{chai2023persistentnature}, as well as learning domain-specific dynamics of abstract scenes~\cite{mahapatra2022controllable}, and very recently general dynamic prior from largescale videos~\cite{li2023generative}.

\noindent \textbf{3D Generative Models.} The recent surge in the quality and diversity of generations orchestrated by 2D image diffusion models poses the question of whether the prior knowledge learned by these models can be used for generating 3D objects and scenes. Indeed, diffusion models have some textual control over the viewpoint. For example, DreamBooth~\cite{dreambooth} shows that the diffusion model can properly react to the words "front", "back", and "side" words the prompt. The seminal work that exploits 2D diffusion for 3D generation, DreamFusion~\cite{dreamfusion}, proposes to optimize NeRF representation by judging the novel view generations with the large-scale pre-trained text-to-image diffusion model~\cite{imagen}. Several follow-up works~\cite{magic3d,fantasia3d} improve the resolution and quality of the resulting 3D assets. On the other hand, Dreambooth3D~\cite{dreambooth3d} introduces additional image control. Although these works can generate reasonable novel views, they require a lengthy optimization process.

Additionally, several works~\cite{richardson2023texture,chen2023text2tex} have proposed to utilize stable diffusion for mesh texturing. TEXTure~\cite{richardson2023texture} uses 2D diffusion models to sequentially in-paint novel regions over the existing mesh, projecting results via a differentiable renderer. Text2Tex~\cite{chen2023text2tex} extends this strategy with an automatic viewpoint-finding approach for optimized re-projection. In both of these works, Stable Diffusion utilizes text as conditioning, which provides only limited control over the generation.

\section{Method}

In this section, we introduce the overall task setup (Sec.~\ref{sec:overview}), the strategy used for generating inputs to the inpainting network (Sec.~\ref{sec:depthsplatting}), three different losses used throughout training (Sec.~\ref{sec:finetuning}), and our inference technique (Sec.~\ref{sec:inference}).    

\subsection{Overview}\label{sec:overview}

\textbf{Novel View Synthesis Task.} Given a single RGB image of a 3D asset (source view) $\sourceview \in \real^{\height \times \width \times 3}$, and the corresponding camera pose $\camerasource \in \real^{3 \times 4}$, we aim to generate a target view $\targetview \in \real^{\height \times \width \times 3}$ of this asset from a novel viewpoint, say $\cameratarget \in \real^{3 \times 4}$. 

\noindent \textbf{Inpainter Inputs.} We start by preparing inputs for the inpainting model, which takes in a partial view of the scene as well as a binary mask that denotes the region to be inpainted. 
We then obtain source view depth $\sourcedepth \in \real^{\height \times \width}$, which is available for our synthetic training set, and can be calculated using an off-the-shelf monocular depth estimator~\cite{bhat2023zoedepth} during inference. \new{Using this depth map, we warp the pixels from the source to the target viewpoint by creating a partial target view  $\targetsplat \in \real^{\height \times \width \times 3}$. We train our \ISD model to inpaint the missing regions of this partial view.} Additionally, we provide the inpainter network with a mask $\masksplat \in \real^{\height \times \width}$ that indicates parts of the image which require inpainting. 

\noindent \textbf{Training and Inference.} We train our \invs model intialized from the Stable Diffusion inpainting checkpoint\footnote{https://huggingface.co/runwayml/stable-diffusion-inpainting} on source-target pairs sampled at random from 20M rendered views of the largescale synthetic {\objaverse}~\cite{deitke2022objaverse} dataset. The finetuning process is outlined in~\cref{sec:finetuning}. Finally, we also modify the DDIM inference which helps to significanly reduce artifacts in the NVS generations, described in ~\cref{sec:inference}.

\begin{figure}[t]
    \centering
    \includegraphics[width=\linewidth]{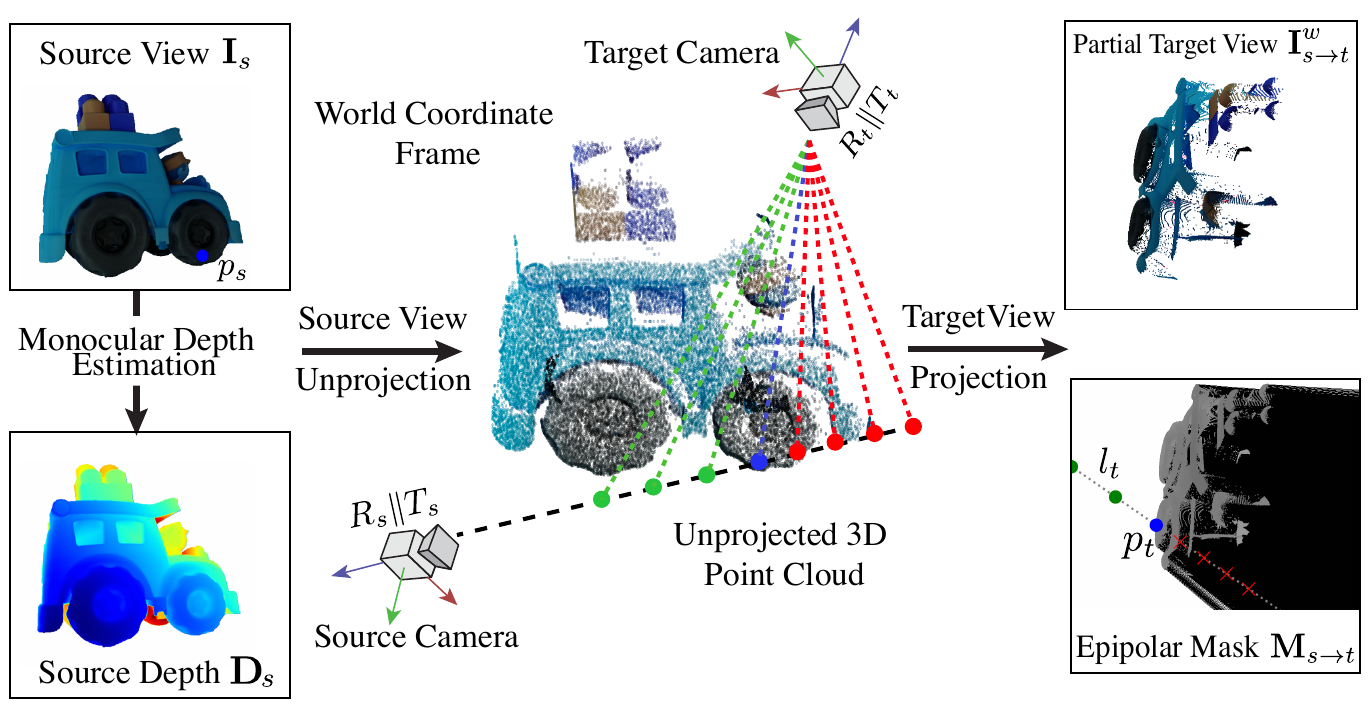}
    \caption{\textbf{Epipolar mask and partial view generation strategy.} Starting with a source view, we use a pre-existing monocular depth estimator to calculate the distance of each point in the image from the camera, creating a depth map. We then use this depth map to "unproject" the 2D image into 3D space, generating a partial point cloud. Next, we take the partial point cloud and "reproject" it onto the target view, essentially projecting the 3D points back onto a 2D image from a different angle. As we do this, we also generate a "visibility mask" which identifies any new areas that become visible in the target view that were not visible in the source view.}
    \label{fig:splating}

\end{figure}
\subsection{Generating Partial View and Epipolar Mask}
\label{sec:depthsplatting}

\begin{figure}[t]
    \centering
    \includegraphics[width=\linewidth]{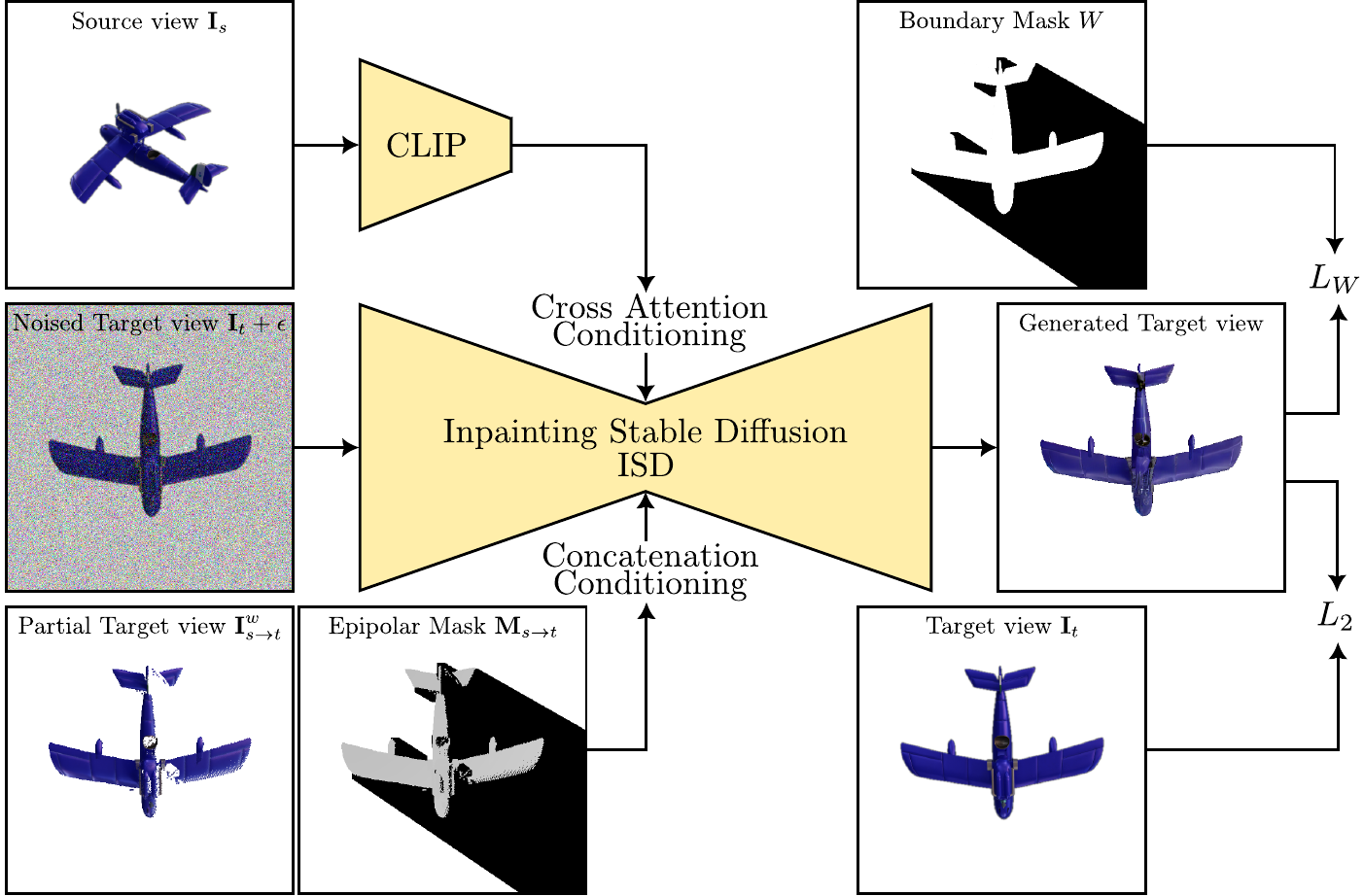}
    \caption{\textbf{Inpainter model training using denoising, and boundary losses.} Inpaining Stable Diffusion (\ISD) accepts a noised target view as well as a partial target view and epipolar mask. All three of these inputs are concatenated before they are fed into the diffusion. Instead of the text condition we use CLIP~\cite{radford2021learning} embedding of the source view, that is provided to the \ISD through cross attention layers. The final generated image is compared against ground truth with $L_2$ loss, moreover to enforce object shape discovery we introduce additional boundary mask loss.}
    \label{fig:diffusion}

\end{figure}

\textbf{Warping source view using depth.} Next, we describe how to unproject the pixels from source view $\sourceview$ into 3D space, and then reproject them into target view $\targetview$ (\aka warping). Let any source pixel from $\sourceview$ be $\pointsource = [x, y, 1]$ defined in homogenous coordinates. We can unproject it into 3D world space by: 
\begin{equation}
    \pointworld = \rotationsource \cdot \depth \cdot \intrinsicsource^{-1} \pointsource + \translationsource,
\end{equation}
where $\intrinsicsource \in \real^{3 \times 3}$ is the source view camera intrinsic, $\camerasource = \camerasourcex \in \real^{3 \times 4}$ is the source camera, and $\depth \in \real$ is the scalar depth value for the point $\pointsource$. \new{Finally, world space point $\pointworld$ can be reprojected in the target view with camera as:  
\begin{equation}
    \pointtarget = \intrinsictarget \cdot \tdepth^{-1} \cdot \rotationtarget^{-1} \cdot (\pointworld - \translationtarget),
\end{equation} 
where $\cameratarget = \cameratargetx$ is the target camera, $\tdepth \in \real$ is scalar target depth, and $\pointtarget$ is target pixel in homogenous coordinates.} Applying the above transform for all foreground pixels in the source view can obtain the partial target view $\targetsplat$. Additionally, when reprojecting points to the target view, we use forward softmax-splatting~\cite{niklaus2020softmax} similar to \cite{li2022infinitenaturezero} to handle overlapping points using z-values. In Figures~\ref{fig:splating} and ~\ref{fig:diffusion}, we visualize the warped target outputs.

\noindent \textbf{How to create inpainting mask?} Reprojecting source pixels to target view only gives us the information about the visible pixels of the object. However, it does not tell the inpainting network anything regarding which regions in the image are newly discovered and which already exist. The simplest method to construct an inpainting mask would be to use all pixels that are not part of the object in the partial target view. However, we find that using such strategy creates a very large inpainting mask, and subsequently \ISD struggles to generalize or maintain consistency with the source view. We show results with this in Sec.~\ref{tab:ablations}. 

\noindent \textbf{Inpainting mask with Epipolar Geometry.} When a light ray falls onto a particular pixel $\pointsource$ in the source view, it corresponds to a line $l_t$ in the target view. This line is known as an epipolar line, as illustrated in Fig.~\ref{fig:splating}. It is worth noting that only a portion of this line in the target view is obstructed, while the rest remains visible. The point $\pointworld$ precisely determines which part is obstructed. Anything preceding $\pointworld$ is visible, whereas anything following it is obstructed. To create the inpainting or visibility mask, we generate rays from each pixel in the source view to the target view until they intersect with the reprojected point. This process yields $\masksplat$, which we refer to as the Epipolar Mask. Fig.~\ref{fig:splating} provides a visual depiction of this procedure along with the resulting mask.

\noindent \new{\textbf{Using smooth inpainting mask using ray angles.}
When creating the epipolar inpainting mask, instead of using a binary value of 0/255 (black/white) at each pixel, we use a smooth value (linearly interpolated) between the source and target camera ray angle at the corresponding 3D world point (projected onto this pixel). Providing this inpainting mask indicates to the ISD \textit{how much camera angle variation has happened at each point} (180 degrees is black, 0 degrees is white). This information is used while training and helps the inpainter ignore/overwrite flipped pixels. Figure~\ref{fig:epipolar} shows a visual example of this.}

\subsection{Training \invs: Inpainter for Novel View Synthesis}
\label{sec:finetuning}
\textbf{Denoising Novel Views.} Equipped with the partial target view and the epipolar mask highlighting regions to be inpainted, we can now train our inpainting model. Concretely, our \ISD takes as conditioning the epipolar mask $\masksplat$, partial target view $\targetsplat$, as well as the \new{CLIP embedding of the source view $\textconditioning$}, and it is trained to denoise a noisy target view $\targetview + \epsilon$. Following previous works~\cite{dhariwal2021diffusion, stable_diffusion}, we utilize epsilon parameterization of diffusion and optimize our network with the following loss:
\new{\begin{equation}
    L_{2} = \left \Vert \epsilon - \ISD \left(\targetview + \epsilon, \masksplat, \targetsplat, \textconditioning \right) \right \Vert^2.
\end{equation}
}
\noindent \textbf{Encoding Source Views with CLIP.}  We replace the text-encoder of CLIP~\cite{radford2021learning} used in the Stable Diffusion model with its image-encoder, to condition generation on source view $\sourceview$. Since the source view does not align well with the target view in RGB image space, we choose to formulate the conditioning via cross-attention instead of using concatenation, unlike previous work~\cite{watson2022novel}.  

\noindent \textbf{Stricter Boundary Loss.} Stable Diffusion inpainting model is trained on real images with diverse backgrounds, and we find that it struggles to generate uniform solid color (white or black) backgrounds. It shows an affinity towards inpainting backgrounds with patterns, or enlarges the object boundaries to cover entirety of inpainting mask.  To tackle this issue, we propose a loss re-weighting that puts more emphasis on target regions where the model has to discover object boundary. Concretely, we introduce re-weighting coefficient $W[\pointtarget] = 1$ if $\pointtarget$ is a pixel that falls within the boundary of known regions, and $W[\pointtarget] = 2$ if $\pointtarget$ is a pixel where the boundary is unknown (as shown in Fig.~\ref{fig:diffusion}). Finally, we obtain the re-weighted loss:
\begin{equation}
    L_{W} = \left \Vert W \left( \epsilon - \ISD(\targetview + \epsilon, \masksplat, \targetsplat, \textconditioning) \right) \right \Vert^2.
\end{equation}

\noindent \textbf{Training on early denoising steps.} We observe that $\ISD$ mostly struggles to decode the shape of the object, while reasonable textures can be obtained even with non-finetuned {\ISD} when using ground truth boundary masks~\cite{chen2023text2tex}. We find that during inference denoising, the shape boundary is discovered much earlier compared to texture, so we additionally fine-tune {\ISD} by sampling noise levels in the first 10\% of the denoiser schedule.

\subsection{Inference}
\label{sec:inference}

\textbf{Rescale and Recenter.} At inference time we wish to generate a novel view of the object from a single source view. Since we do not have ground truth source view depth we rely on monocular depth predictor ZoeDepth~\cite{bhat2023zoedepth}. However, depth estimators can predict object depth only up to unknown scale, hence, we recenter the projected world points to the origin and rescale them into a cube, which follows the setup used in rendering our dataset. 

\noindent \textbf{Guiding Inference using Partial Target View.} We observe that instead of starting the backward denoising process from pure noise, we can significantly boost the quality of the generated views by starting with a noisy version of the image $\targetsplat$ (see~\cref{sec:ablation}).

\section{Experiments}

In this section, we provide details on our training and evaluation datasets (Sec.~\ref{sec:train_dataset}, Sec.~\ref{sec:eval_dataset}), compare our method to three NVS models (Sec.~\ref{sec:compare_sota}), and provide an ablation study (Sec.~\ref{sec:ablation}) of each component.

\subsection{Training Setup.}
\label{sec:train_dataset}

\textbf{Objaverse and Rendering Setup.} To train \invs we require paired data consisting of source and target views. To generate this data, we utilize the extensive {\objaverse} dataset, which contains nearly 800,000 3D assets. We employ Blender as our rendering software and begin by recentering all scenes at the origin. Additionally, we rescale the bounding box of each scene to fit within a $[-1, 1]^3$ cube. For each object in the dataset, we randomly generate 24 camera viewpoints within predefined boundaries. The radius of the viewpoint is sampled from a range of $3$ to $4$, and the field of view (FoV) is set to 50 degrees. Using these viewpoints, we render both the images and corresponding depth maps. We utilize the Cycles engine in Blender and employ 128 samples per ray for rendering. All images are rendered at a resolution of $512\times512$ pixels.

\noindent \textbf{Selecting good camera poses.} To add diversity to the lighting conditions, we randomly sample lighting from a collection of 100 environmental maps. These maps provide a range of indoor and outdoor lighting conditions with varying intensity levels. It's worth noting that in the Objaverse dataset, most assets are oriented with the Z-axis pointing upwards. Consequently, synthesizing the object from extreme bottom or top view angles can be challenging. \new{For instance, when objects are placed on a platform, viewing them from below makes it almost impossible to accurately determine the opposite view without additional information.} To address this issue, we empirically determine that sampling the polar angle $\theta$ from a uniform distribution between -65 and 75 degrees provides reasonably accurate views on average. We sample the azimuth angle $\phi$ randomly between 0 and 360 degrees.

By utilizing all 800,000 assets in the Objaverse dataset, we render a total of 19 million images to train our \invs model.

\noindent \textbf{Model and Training details.} We performed fine-tuning on the pretrained Inpainting Stable Diffusion (\ISD) v1.5 checkpoint to adapt it for our task. This model is capable of generating high-resolution images with dimensions of $512\times512$, utilizing a latent space of dimensions $64\times64\times4$. During the fine-tuning process, we employed a sequential training approach, consisting of three stages with separate losses previously introduced: a) denoising, b) boundary loss, and c) early steps training. Our final model was trained on 96 A100 GPUs with each stage training over 7 days.
\begin{table}[t]
    \caption{\new{Comparison with baselines on \textbf{Google Scanned Objects} dataset.}}  

    \begin{center}
		\setlength{\tabcolsep}{2pt}
		\begin{tabular}{ll cc cc c}
			\toprule
            \multirow{2}{*}{\texttt{\#}} & \multirow{2}{*}{\textbf{{Method}}} & \multicolumn{2}{c}{\textbf{PSNR} $\uparrow$} & \multicolumn{2}{c}{\textbf{SSIM} $\uparrow$} & \multirow{2}{*}{\textbf{LPIPS} $\downarrow$}  
            \\
			\cmidrule(lr){3-4} \cmidrule(lr){5-6} 
             &  & {mask} &  {unmask} &  {mask} & {unmask} & 
            \\
            \midrule
            \texttt{1} & \pointe & 8.90 & 12.04 & 0.18 & \underline{0.82} & \underline{0.25}  
            \\ 
            \texttt{2} & \shapee & 10.39 & 12.18 & \underline{0.30} & \underline{0.82} & 0.29   
            \\ 
            \texttt{3} & \zeroone  & 14.74 & \underline{14.70} & \textbf{0.34} & \textbf{0.84} & \underline{0.25}    

            \\ 
            \cmidrule(lr){0-6}
            \texttt{4} & \emph{Original \ISD} & \underline{15.03} & 13.25 & 0.09 & 0.49 & 0.38 
           \\
            \texttt{5} & \invs (ours) & \textbf{18.95} & \textbf{19.83} & \underline{0.30} & 0.80 & \textbf{0.24} 
            \\ 
            \bottomrule
		\end{tabular}
	\end{center}
    \label{tab:gso}
\end{table}

\begin{table}[t]
    \caption{\new{Comparison with baselines on \textbf{Ray-traced Multiview} data.}} 

    \begin{center}
		\setlength{\tabcolsep}{2pt}
		\begin{tabular}{ll cc cc c}
			\toprule
            \multirow{2}{*}{\texttt{\#}} & \multirow{2}{*}{\textbf{{Method}}} & \multicolumn{2}{c}{\textbf{PSNR} $\uparrow$} & \multicolumn{2}{c}{\textbf{SSIM} $\uparrow$} & \multirow{2}{*}{\textbf{LPIPS} $\downarrow$} 
            \\
			\cmidrule(lr){3-4} \cmidrule(lr){5-6} 
             &  & {mask} &  {unmask} &  {mask} & {unmask} &  
            \\
            \midrule
            \texttt{1} & \pointe & 7.40 & 10.44 & 0.14 &\textbf{ 0.67} & \textbf{0.41}  
            \\ 
            \texttt{2} & \shapee & 8.35 & 9.74 & \textbf{0.17} & \underline{0.65} & \underline{0.48}   
            \\ 
            \texttt{3} & \zeroone  & 9.09 & 8.29 & \underline{0.16} & 0.58 & 0.50    

            \\ 
            \cmidrule(lr){0-6}

            \texttt{4} & \emph{Original \ISD} & \underline{14.61} & \underline{11.25} & 0.09 & 0.27 & 0.65 

           \\
            \texttt{5} & \invs (ours) & \textbf{16.83} & \textbf{17.82} & 0.09 & 0.5 & 0.49 
            \\ 
            \bottomrule
		\end{tabular}
	\end{center}
    \label{tab:rtmv}

\end{table}

\begin{table}[t]
    \caption{\new{Comparison with baselines on \textbf{Common Objects in 3D} dataset.}} 

    \begin{center}
		\setlength{\tabcolsep}{2pt}
		\begin{tabular}{ll cc cc c}
			\toprule
            \multirow{2}{*}{\texttt{\#}} & \multirow{2}{*}{\textbf{{Method}}} & \multicolumn{2}{c}{\textbf{PSNR} $\uparrow$} & \multicolumn{2}{c}{\textbf{SSIM} $\uparrow$} & \multirow{2}{*}{\textbf{LPIPS} $\downarrow$}
            \\
			\cmidrule(lr){3-4} \cmidrule(lr){5-6} 
             &  & {mask} &  {unmask} &  {mask} & {unmask} &   %
            \\
            \midrule
            \texttt{1} & \pointe & 9.37 & 10.10 & 0.22 & \underline{0.72} & \underline{0.38}  
            \\ 
            \texttt{2} & \shapee & 10.67 & 10.01 & \textbf{0.33} & \textbf{0.73} & 0.42   
            \\ 
            \texttt{3} & \zeroone  & 12.32 & 9.91 & \textbf{0.33} & 0.69 & 0.42    

            \\ 
            \cmidrule(lr){0-6}

            \texttt{4} & \emph{Original \ISD} & \underline{16.43} & \underline{13.56} & \underline{0.24} & 0.46 & 0.44 

           \\
            \texttt{5} & \invs (ours) & \textbf{17.58} & \textbf{17.39} &\textbf{ 0.33} & 0.65 &\textbf{ 0.36} 
            \\ 
            \bottomrule
		\end{tabular}
	\end{center}
    \label{tab:co3d}

\end{table}

\begin{figure*}[t]
    \centering
    \includegraphics[width=\linewidth]{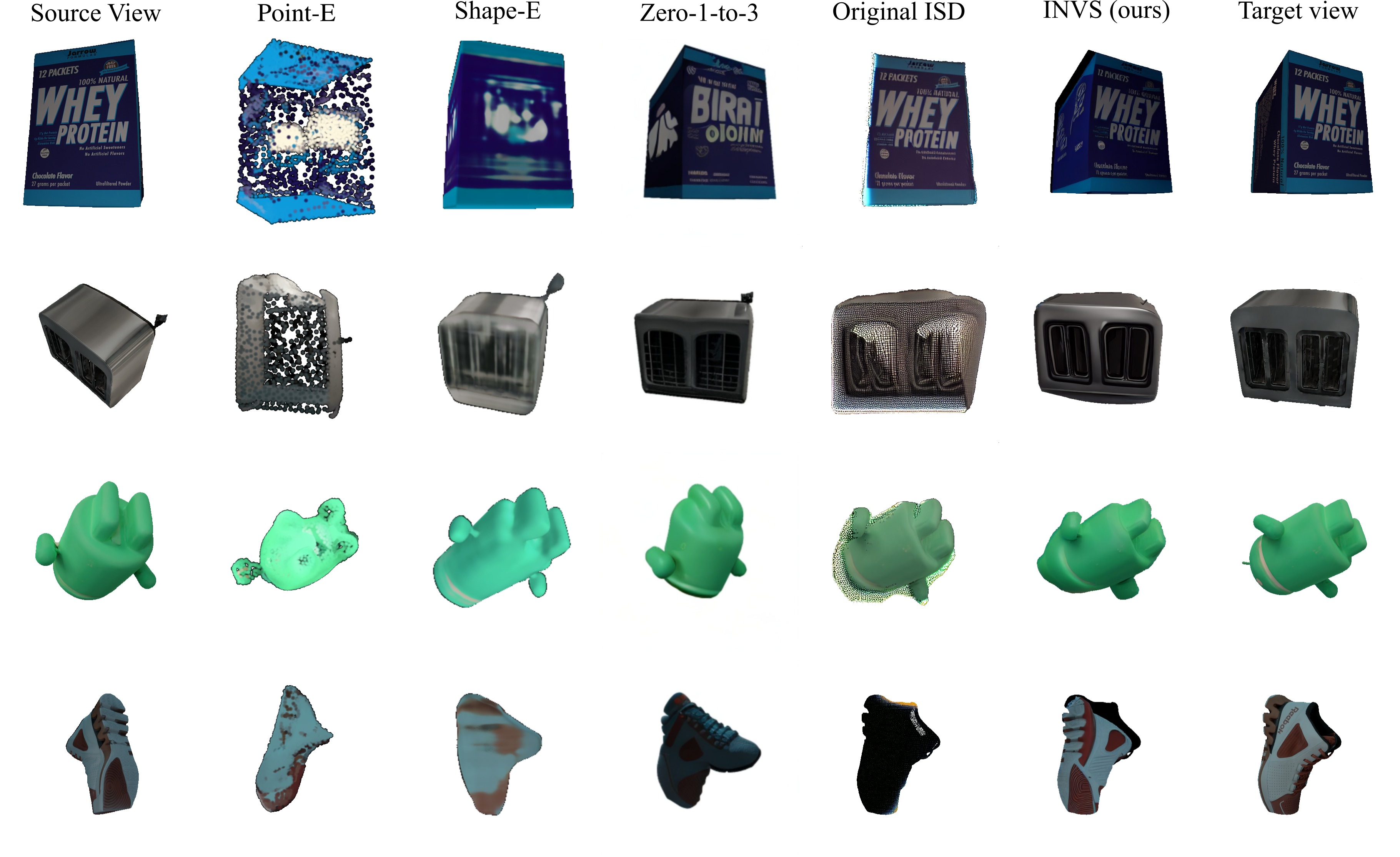}
    \caption{\textbf{\new{Comparison with SoTA methods on NVS task (GSO dataset).}} The first column is input, columns two to four are baselines: \pointe~\cite{nichol2022pointe}, \shapee~\cite{jun2023shape} and \zeroone~\cite{zero1to3}. Fifth column is untrained \ISD, and last two columns is \invs and ground truth.}
    \label{fig:sota}
\end{figure*}
\begin{figure}[t]
    \centering
    \includegraphics[width=\linewidth]{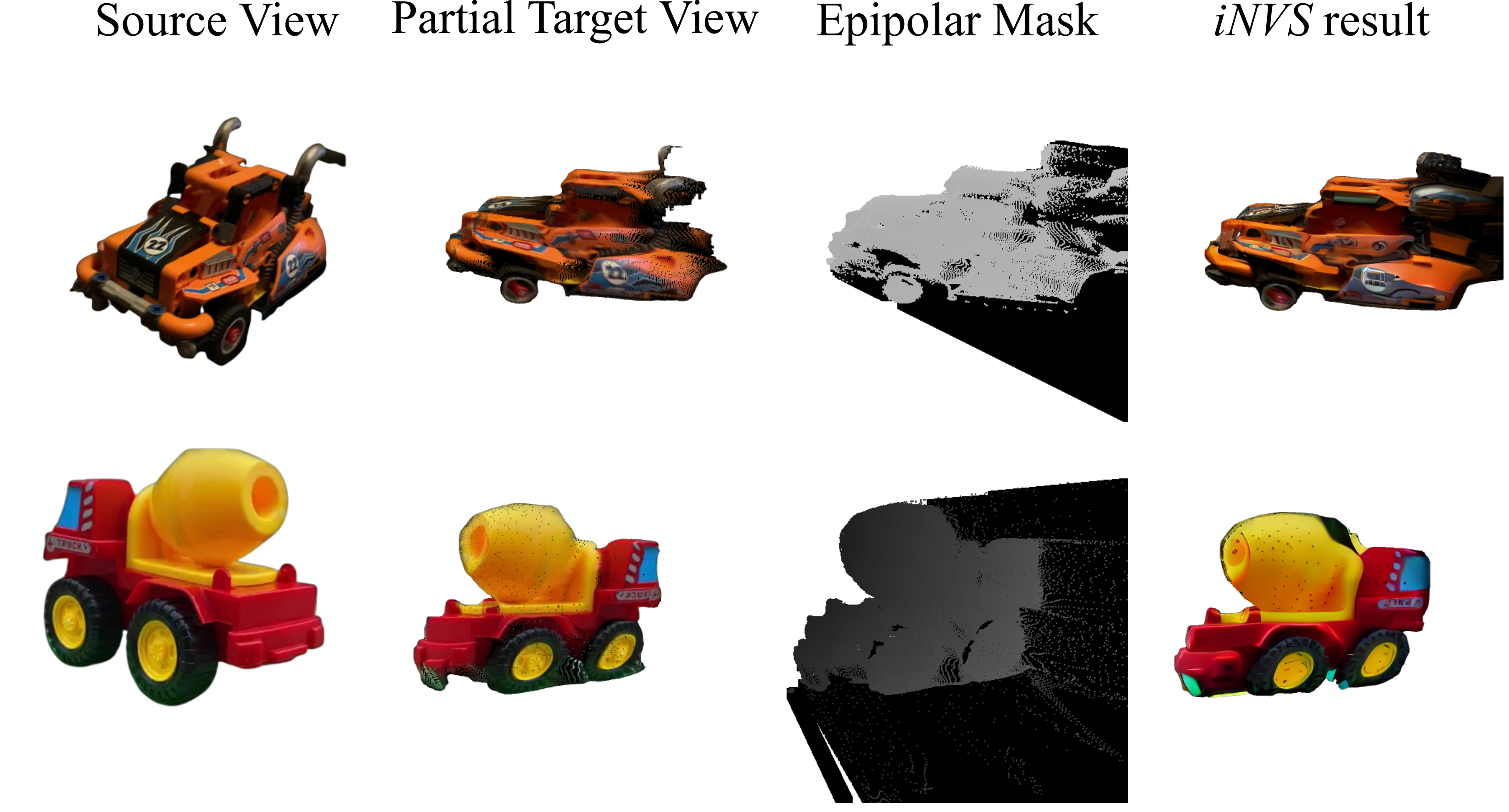}
     \caption{\textbf{Partial target view and epipolar masks on CO3D dataset.} The first column shows input. Second column show the warped source view. 
     \new{The third column demonstrates the smooth inpainting epipolar mask. Notice that the inpainting mask for cement dump truck (second row) is much darker compared to the race car (first row) due to larger angle variation (details in~\cref{sec:depthsplatting}).} 
     Last column shows generated result.}
    \label{fig:epipolar}

\end{figure}

\begin{figure*}[t]
    \centering
    \includegraphics[width=\linewidth]{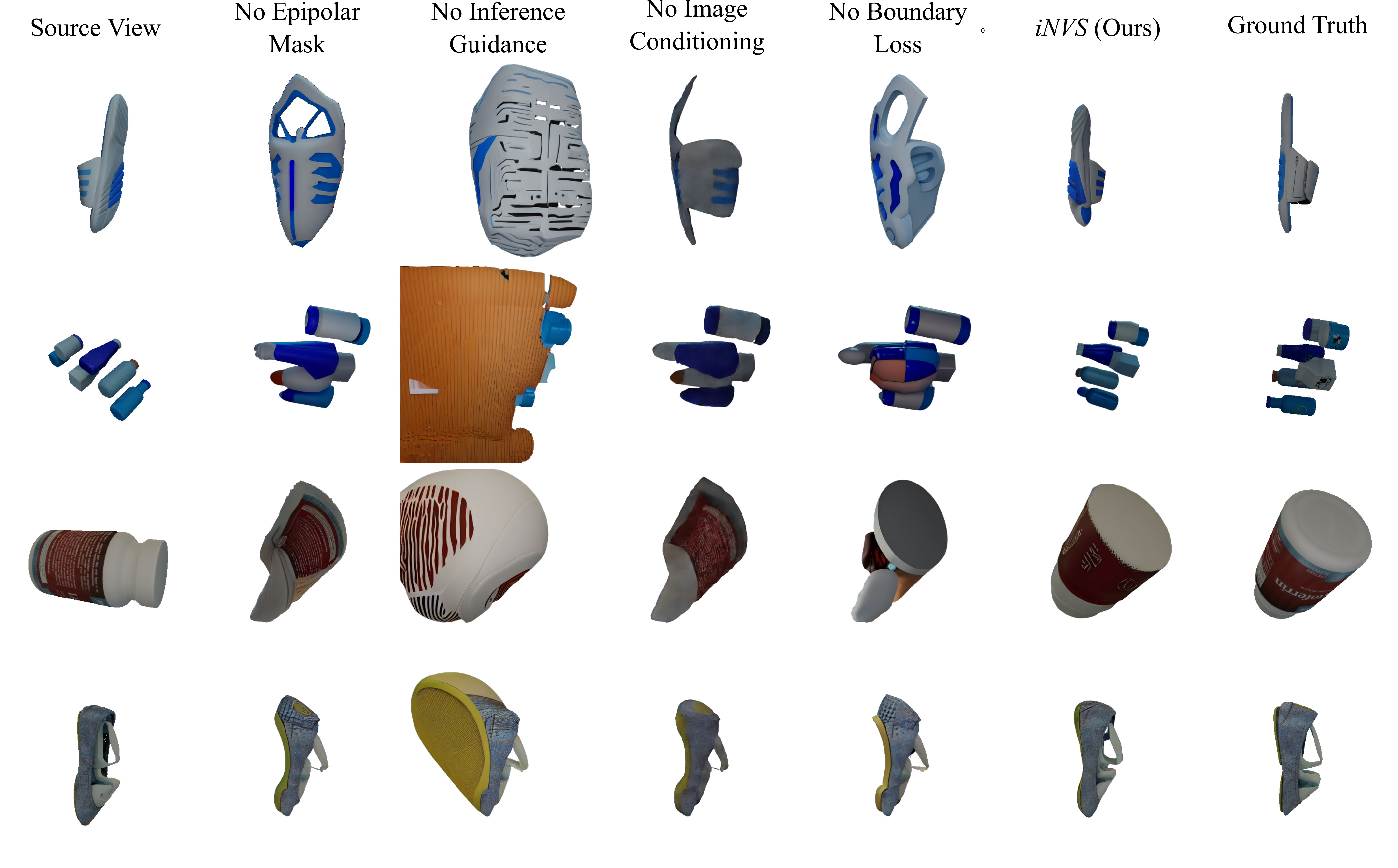}
    \caption{\textbf{Ablation Study of \invs on GSO dataset.} We describe each column from left to right. First, we show input image. Second, we show generation without the epipolar mask. Third, we show generation without inference guidance (Section~\ref{sec:inference}). Fourth, we skip source-view conditioning via CLIP during inference. Fifth, we show a variant of our method without boundary loss $L_W$. The last two columns show result from \invs and ground truth.}
    \label{fig:ablations}
\end{figure*}

\subsection{Evaluation}
\label{sec:eval_dataset}

\textbf{Datasets.} We evaluate how well \invs generalizes at generating novel target views across three different datasets. Specifically, we use two synthetics datasets, \emph{Google Scanned Objects} (\gso)~\cite{downs2022google, }, and \emph{Ray-Traced Multi-View} (\rtmv)~\cite{tremblay2022rtmv}. The \gso dataset contains nearly one thousand photorealistic 3D models, which we render using Blender following the same setup used for generating training data (described in \ref{sec:train_dataset}). The \rtmv dataset contains high quality renderings of nearly 2000 scenes from 4 different sources, and we filter out the scenes that contain \gso objects. Finally, we also evaluate on real videos from the \emph{Common Objects in 3D} {\cothreed}~\cite{reizenstein2021common} dataset, which is a dataset of 19,000 videos of common objects spanning 50 categories. 

\noindent \textbf{Metrics.} Following prior work~\cite{zero1to3}, we compare \invs and baselines with three different metrics covering different aspects of image similarity: PSNR, SSIM~\cite{ssim} and LPIPS~\cite{zhang2018unreasonable}. During evaluation, for every object (or scene) we sample two random views $\sourceview$ and $\targetview$ along with their camera poses $\camerasource$ and $\cameratarget$. Next, starting from $\sourceview$ we can compute relative camera transformation and generate a target view. 

\noindent \new{\textbf{Masked Metrics.} For PSNR and SSIM metrics we find that filtering out background (using the ground truth mask) before comparison helps to avoid spurious gains, hence we report masked metrics.}

\subsection{Baseline Comparisons}
\label{sec:compare_sota}

\textbf{\zeroone~\cite{zero1to3}} is an image and camera pose conditioned diffusion model which leveraged a pretrained Stable Diffusion called Image Variations~\cite{lambda-labs} and finetuned it on Objaverse renders. Unlike our method, Zero-1-to-3 is trained to generate novel views from scratch, and is prone to cause inconsistency between source and target views (see results for more details). We use the official codebase\footnote{https://github.com/cvlab-columbia/zero123} and checkpoints with our datasets for evaluation.

\noindent \textbf{\pointe~\cite{nichol2022pointe}} is an image-conditioned diffusion model which operates over 3D point clouds to generate objects. We use the official codebase and checkpoint\footnote{https://github.com/openai/point-e}, and use the settings mentioned in the paper to generate point clouds with 4,000 points. We render the point cloud from the target viewpoints for novel views. 

\noindent \textbf{{\shapee}~\cite{jun2023shape}} is a conditional generative model for 3D, which directly outputs the parameters of implicit functions that can be rendered directly as neural radiance fields. It is a two-stage model that involves generating a latent code for each 3D asset and then uses a diffusion model to denoise this latent code. We use the official codebase and checkpoint\footnote{https://github.com/openai/shap-e}, and render outputs as neural radiance field from the target viewpoint. 

\noindent \textbf{Our method achieves good PSNR and comparable LPIPS.} We find that our method achieves the higher PSNR and comparable LPIPS scores compared to all other baselines on the Google Scanned Objects (\gso) and Common Objects in 3D (\cothreed) benchmarks (Table~\ref{tab:gso} and~\ref{tab:rtmv}). This indicates that our method performs well in terms of noise reduction and perceptual similarity in both synthetic and real data scenarios. The \cothreed dataset consists of real-world views captured from free-form videos, while the \gso dataset contains virtual scans of 3D photorealistic assets. On the Ray-traced Multiview (\rtmv) dataset, we find that our method is able to outperform all baselines in PSNR, but falls short on the SSIM and LPIPS metrics (shown in Table~\ref{tab:rtmv}). We attribute this low performance to out-of-distribution variations in lighting across our rendering setup (described in Section~\ref{sec:train_dataset} compared to \rtmv). 

\noindent \textbf{Structural Similarity is compromised in generated views.} It is worth mentioning that our method consistently underperforms on the SSIM metric across all datasets. We find that this occurs primarily due to misalignment in monocular depth estimator. We observe that under significant viewpoint variations, the monocular depth estimator fails to generate consistent depth across different parts of the objects. This inconsistency leads to distortions in the generated images and lower SSIM scores \new{(see Section~\ref{ssec:limits})}.

\new{\noindent \textbf{Masked metrics help disambiguate performance across baselines.} We notice that Shap-E and Point-E often produce tiny objects, and their white background pixels (matching the target) lead to majority of their unmasked gains, thus outperforming Zero123 quantitatively. However, using masked metrics we notice these trends change (see PSNR and SSIM in Tables~\ref{tab:rtmv} and~\ref{tab:co3d}).}

\subsection{Qualitative Results}

We present the visualization of the results obtained from our method and the baselines in Figure~\ref{fig:sota} and draw the following conclusions:

\noindent \textbf{Preservation of Text and Fine Details.} We employ monocular depth to unproject pixels into 3D space and warp them to the target viewpoint. This technique enables us to preserve a significant amount of textual information and finer details across multiple viewpoints. This is clearly demonstrated in Row 1, where our method successfully retains the text "Whey Protein" between the source and target views, unlike the \zeroone baseline. Additionally, our method shows less distortion in the structural details of objects, as seen in Row 2, where \zeroone alters the toaster oven into grills, while our method accurately preserves this detail. The benefits of this technique is further highlighted in~\cref{fig:epipolar}, where we show that for many viewpoints most of the partial target view can be directly reused to significantly simplify the job of inpainting network. 

\begin{figure*}[t]
     \includegraphics[width=0.975\linewidth]{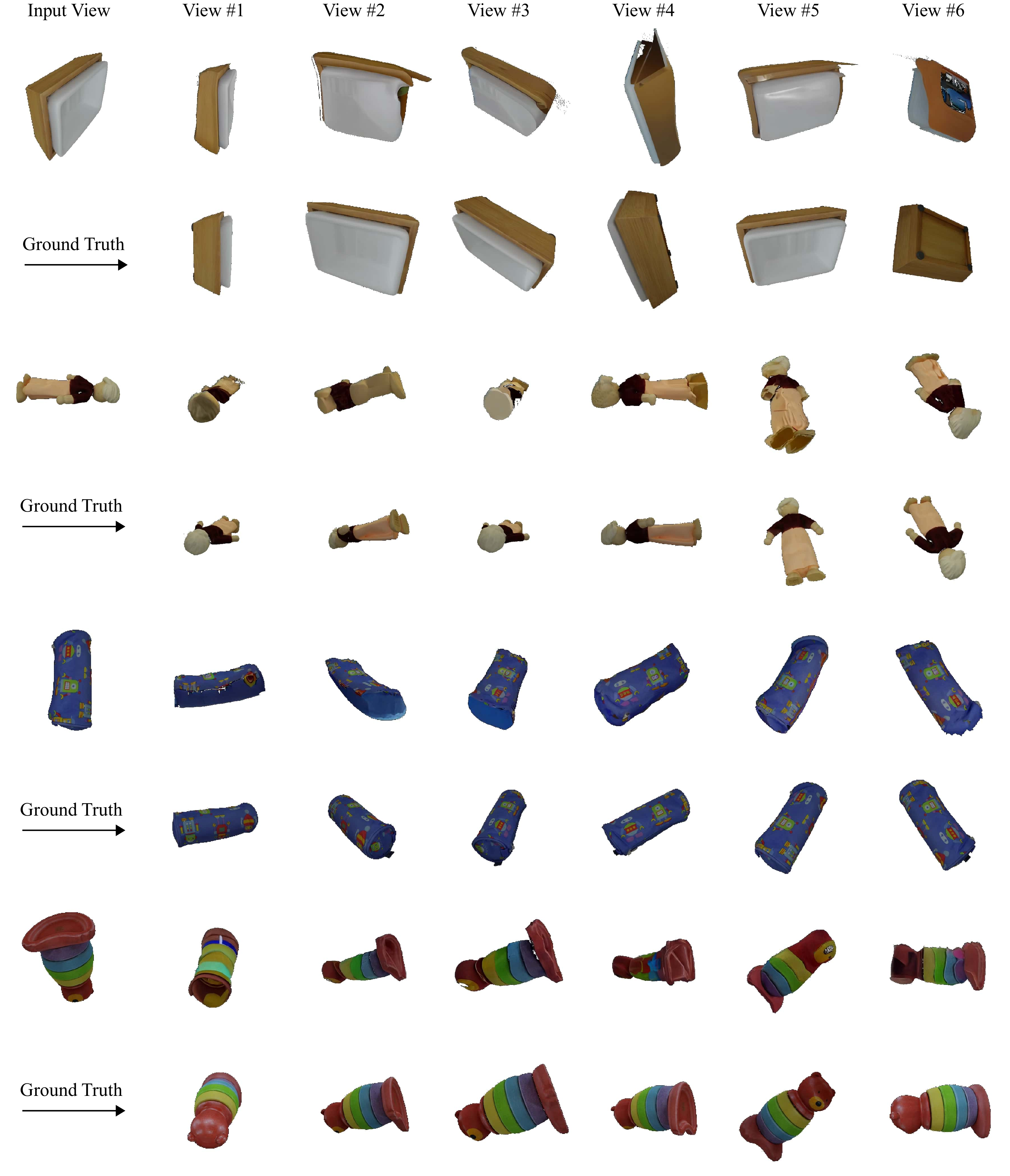}
    \caption{\new{\textbf{Multiple novel views from single image.} We show six randomly sampled camera views given an input image, and corresponding ground truth.}}
    \label{fig:view2}

\end{figure*}

\noindent \new{\textbf{Consistent synthesis for single object across multiple views.} In Figures~\ref{fig:view1} and ~\ref{fig:view2}, we show NVS results across many different objects from a fixed source view, and randomly selecting six target views. We find our generations to be largely coherent across viewpoints, and more stochastic under large viewpoint variations.}

\noindent \textbf{Faithful Compliance with Viewpoint Variations:} Another challenge in novel view synthesis is maintaining control over the generated views, particularly when dealing with significant viewpoint variations. Rows 3 and 4 exemplify this issue, showing that \zeroone struggles to exercise accurate control over the generated images, resulting in the android and the shoe being created from incorrect viewpoints.

\noindent \textbf{Outperforming 3D Diffusion Models:} Our method surpasses the performance of the \pointe and \shapee baselines, both of which are variants of 3D diffusion models. We believe that utilization of a large, pretrained inpainting model in our method contributes to the generation of visually superior results.

\subsection{Ablation Study}
\label{sec:ablation}

We conduct a series of ablation experiments and analysis to evaluate the effectiveness of our proposed changes and additional training. The quantitative performance results for all metrics are reported in Table~\ref{tab:ablations}, and visuals are presented in Figure~\ref{fig:ablations}.

\noindent \textbf{Epipolar Mask helps to constrain NVS generation.} The epipolar mask allows us to control the extent of the inpainting necessary in the warped image, as depicted in Figures~\ref{fig:splating} and~\ref{fig:diffusion}. When we omit the use of this mask, the inpainting model faces challenges in understanding the relative orientation between the source view and the target view. As a result, the performance of the model is compromised, leading to distorted and exaggerated generated images.

\noindent \textbf{Guiding denoising inference with partial target view yields significant benefits.} We find that during the 500 step DDIM inference \invs generates a rough outline of the target view within the first 100 steps. Given that we have a partial target view available after warping the source pixels to the target viewpoint, we utilize it as guidance. Specifically, we replace the output of the first 10 DDIM steps with a noised version of the partial target view, similar to Repaint~\cite{lugmayr2022repaint}. This approach proves to be immensely helpful in preventing the inpainting model from generating arbitrary boundaries, shown in Figure~\ref{fig:ablations}, third column.

\noindent \textbf{Without image conditioning our model is unable exploit source view information well.} When there is a significant variation in viewpoint between the target and source images, the warped image becomes less informative for the inpainting process. In such scenarios, it becomes essential for the inpainting model to heavily rely on the source view and the learned 3D priors for accurate autocompletion. By removing the conditioning on the source view, the model's ability to generate high-quality and consistent views is compromised, as demonstrated in Figure~\ref{fig:ablations}, fourth column.

\noindent \textbf{Boundary loss enhances generations.} Furthermore, we find that enhancing the boundary loss improves the generation quality leading to consistent visuals, evident from Figure~\ref{fig:ablations}, fifth column.

\begin{table}[t]
    \caption{\textbf{Ablation Study.} First row is our method \invs. The second row shows our method where the epipolar mask is replaced with a full mask that covers all non-splatted pixels. The third row is our method without using a partial target view as diffusion input in the first steps as guidance. The fourth row is our method without conditioning on the source image. The fifth row demonstrates our method without $L_W$. Finally we highlight our model with the original training schedule. Also for the reference we show the performance of the original unaltered \ISD.}

    \begin{tabular}{lccc}
        \toprule
        \textbf{Method} & \textbf{PSNR} $\uparrow$ & \textbf{SSIM} $\uparrow$ & \textbf{LPIPS} $\downarrow$ \\
        \cmidrule(lr){1-4}
        \invs \emph{(ours)} & \textbf{19.83} & \textbf{0.8} & \textbf{0.24} \\
        
        {\emph{- epipolar mask}} & $17.39_{\color{red}-2.44}$ & $0.65_{\color{red}-0.15}$ & $0.36_{\color{red}+0.12}$ \\
        
        {\emph{- inference guidance}} & $17.48_{\color{red}-2.35}$ & $0.70_{\color{red}-0.1}$ & $0.31_{\color{red}+0.07}$ \\
        
        {\emph{- image conditioning}} & $16.57_{\color{red}-3.26}$ & $0.70_{\color{red}-0.01}$ & $0.30_{\color{red}+0.06}$ \\
        
        {\emph{- boundary loss}} & $19.10_{\color{red}-0.73}$ & $0.77_{\color{red}-0.03}$ & $0.27_{\color{red}+0.03}$ \\
        
        {\emph{- early steps training}} & $19.70_{\color{red}-0.13}$ & $0.78_{\color{red}-0.03}$ & $0.26_{\color{red}+0.02}$ \\
        
        \cmidrule(lr){1-4}
        
        \emph{Original \ISD} & $13.25_{\color{red}-6.58}$ & $0.49_{\color{red}-0.31}$ & $0.38_{\color{red}+0.14}$  \\
        \bottomrule 
    \end{tabular}
    \label{tab:ablations}

\end{table}

\begin{figure*}[t]
     \includegraphics[width=0.95\linewidth]{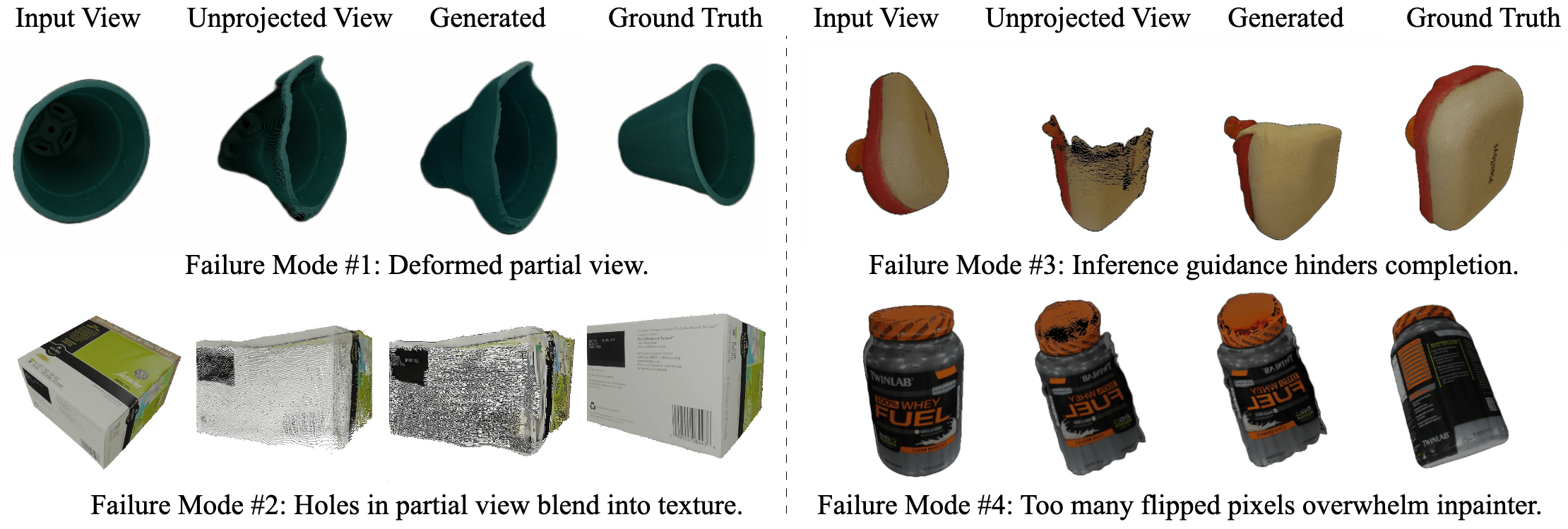}
    \caption{\new{\textbf{Failure Modes.} \textbf{Left:} Imperfect depth maps cause issues in structure and texture. \textbf{Right:} Inference time tricks can occasionally hinder generations. }}
    \label{fig:fail}

     \includegraphics[width=0.925\linewidth]{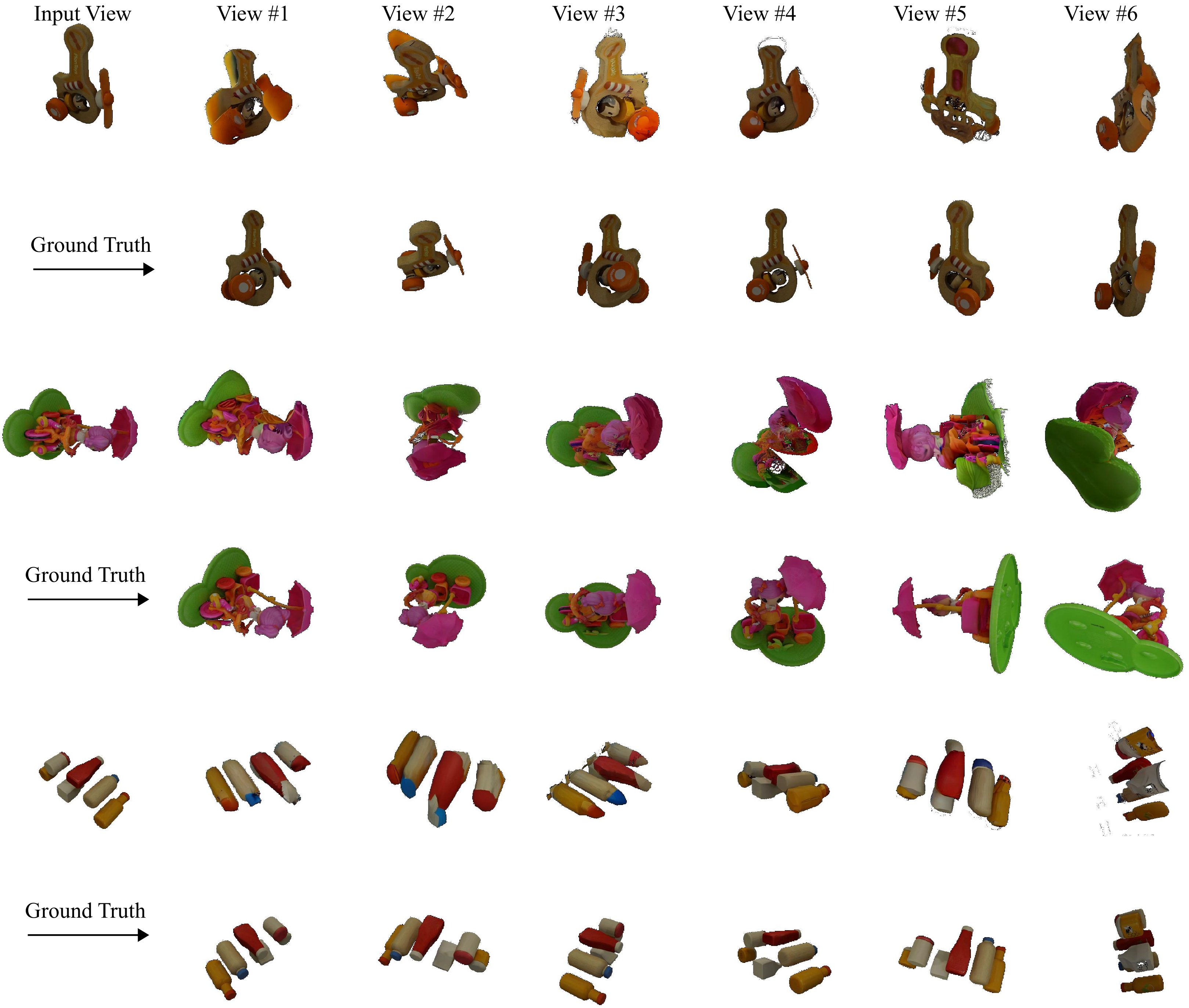}
    \caption{\new{\textbf{Multiple novel views from single image.} We show six randomly sampled camera views given an input image, and corresponding ground truth.}}
    \label{fig:view1}
\end{figure*}

\new{
\subsection{Current Limitations}
\label{ssec:limits}
\noindent \textbf{Imprecise monocular depth can lead to structure and texture problems.} We notice that ZoeDepth~\cite{bhat2023zoedepth} can generate depth maps that distorts flat surfaces, which leads to unrealistically deformed surfaces or incorrect texture predictions. We put visuals of these in failure modes \#1 and \#2 in Figure~\ref{fig:fail}. 

\noindent \textbf{Inference guidance can occasionally lead to incomplete or flipped output images.} Recall from Section~\ref{sec:inference} that we use the noisy partial target view for the first 10 DDIM steps. However, since this view is incomplete and may contain flipped pixels (under high camera variations), this trick occasionally creates incomplete outputs or lead to reversed-view generated images. We put visuals of these in failure modes \#3 and \#4 in Figure~\ref{fig:fail}. 

}

\section{Conclusion}
In this work we propose an approach for novel view synthesis that provides a significant advancement in terms of quality of the results and the coverage of the different object categories. 
Our approach combines recent advancement of diffusion models with epipolar geometry. By training on a large scale \objaverse dataset we were able to re-purpose Inpainting Stable Diffusion for the novel view synthesis task. Surprisingly we found that after finetuning our method gained an understanding of the underlying shape, even for the objects that were not seen during the training.
Our approach demonstrates sizable improvement over state-of-the-art novel view synthesis methods, especially when considering texture preservation from the input image.
One limitation of our approach is inability to generate consistent textures in the regions that are not visible in the original image. 
This however opens the exciting opportunity for future research, which can explore auto-regressive schemes of novel view generation. 

\begin{acks}
We thank Ziyi Wu for helping with aligning visuals of Point-E and Shap-E baselines, as well as organizing cited works. We also thank Colin Eles for helping with infrastructure required for largescale training, and Pratiksha Bhattasamant for reviewing early drafts of this work. 
Finally, we would like to thank the Siggraph Asia reviewing committee for their invaluable feedback.

\end{acks}

\clearpage
\bibliographystyle{ACM-Reference-Format}
\bibliography{refs}

\clearpage
\appendix
\section{Training Hyperparameters}
\begin{table}[h!]
\center
\begin{tabular}{clc}
\toprule
    \# & Hyperparameters & Value \\
    \midrule
    {1} & Base Learning rate & 1e-5 \\
    {2} & Learning rate decay & N/A \\
    {3} & Loss Type & L1 \\ 
    {4} & Source-view / Inpaint Mask dropout & 0.05 \\
    {5} & Classifier-free guidance & 9.0 \\
    {6} & Effective batch size & 1152 \\
    {7} & Effective batch size & 1152 \\
    {8} & DDIM Steps & 300/500 \\
    {9} & Partial View Guidance Steps & 10 \\
    {10} & Boundary Loss Weight & 2.0 \\
    {11} & CLIP Frozen & False \\
    {12} & Renders background color & Black \\
    {13} & Image Resolution & 512 \\
    {14} & Learning rate linear warmup & 100 steps \\ 
    \midrule
\end{tabular}
\caption{Hyperparameter choices for \invs.} 
\label{tab:hyperparameters_supp}
\end{table}

\section{Experiment with Diffusion-based Monocular Depth Network}
\textbf{Idea and Setup.} Since incorrect predictions from monocular depth network lead to failure cases, we trained a separate diffusion-based monocular depth prediction network on the Objaverse dataset. For this, we simply inflate the first convolution layer of Stable Diffusion~\cite{stable_diffusion} to take input RGB image as additional conditioning via concatenation. The model is trained with L1 loss to generate monocular depths, and it is initialized with pretrained text-to-image Stable Diffusion weights. 

\noindent \textbf{Result.} We found that the compression in latent-space of diffusion model due to the VAE caused issues in the predicted depth maps. This in turn led to \textit{jittery outputs and reprojections} leading to subpar performance. Since we trained only on Objaverse dataset, we did not require explicit rescaling and recentering when using predicted depth. We believe improvement in this setup could be achieved by retraining the Autoencoders to reconstruct depth maps, similar to~\cite{stan2023ldm3d}. \textit{We put visuals of this experiment in supplementary video.}

\end{document}